\documentclass[twoside,leqno,twocolumn]{article}

\usepackage[letterpaper]{geometry}

\usepackage{ltexpprt}
\usepackage{hyperref}

\usepackage{diagbox}
\usepackage{multirow}
\usepackage{multicol}
\usepackage{algorithm}
\usepackage{enumitem}
\usepackage{algorithmic}
\usepackage{booktabs}
\usepackage{graphicx}
\usepackage{subfigure}
\usepackage{amsfonts}

\begin{document}

\title{Context-aware Domain Adaptation for Time Series Anomaly Detection}
\author{
Kwei-Herng Lai\thanks{Rice University. \{khlai,Kaixiong.Zhou,Xia.Hu\}@rice.edu}
\and Lan Wang\thanks{Visa Research. \{lwang4,hchen,feiwang,haoyang\}@visa.com} \and Huiyuan Chen\footnotemark[2] \and Kaixiong Zhou\footnotemark[1] \and Fei Wang\footnotemark[2] \and Hao Yang\footnotemark[2] \and Xia Hu\footnotemark[1]
}

\date{}

\maketitle


\fancyfoot[R]{\scriptsize{Copyright \textcopyright\ 2023 by SIAM\\
Unauthorized reproduction of this article is prohibited}}





\begin{abstract}
Time series anomaly detection is a challenging task with a wide range of real-world applications. Due to label sparsity, training a deep anomaly detector often relies on unsupervised approaches. Recent efforts have been devoted to time series domain adaptation to leverage knowledge from similar domains. However, existing solutions may suffer from negative knowledge transfer on anomalies due to their diversity and sparsity. Motivated by the empirical study of context alignment between two domains, we aim to transfer knowledge between two domains via adaptively sampling context information for two domains. This is challenging because it requires simultaneously modeling the complex in-domain temporal dependencies and cross-domain correlations while exploiting label information from the source domain. To this end, we propose a framework that combines context sampling and anomaly detection into a joint learning procedure. We formulate context sampling into the Markov decision process and exploit deep reinforcement learning to optimize the time series domain adaptation process via context sampling and design a tailored reward function to generate domain-invariant features that better align two domains for anomaly detection. Experiments on three public datasets show promise for knowledge transfer between two similar domains and two entirely different domains.

\end{abstract}

\vspace{-10pt}
\section{Introduction}
Detecting anomalies from time series data has a wide variety of applications~\cite{kim2020ai,maceachern2020configurable,al2021financial,lai2021revisiting,lai2021tods,wang2021forecast,zha2020meta,li2020pyodds} in various domains and is very challenging due to the limited access to the label information and complex dependencies between individual time points. Nevertheless, modeling time series data with limited label information leads to sub-optimal performance, and therefore increasing research efforts are devoted to time series domain adaptation. There are two types of existing approaches:  domain discrepancy minimization and domain discrimination. The former one~\cite{cai2021time,du2021adarnn,long2013transfer}, which forcefully minimizes metric distances between the mapped subsequences of two domains in a shared subspace, may lead to a sub-optimal shared subspace due to the  distinction in the data distribution of two domains. The latter one~\cite{jin2021attention,du2021adarnn,tzeng2017adversarial}, which extracts domain invariant features from subsequences of two domains with adversarial learning, may be intractable when the lengths of subsequences from two domains are not well-aligned.

\begin{figure}
    \centering
    \includegraphics[width=0.95\linewidth]{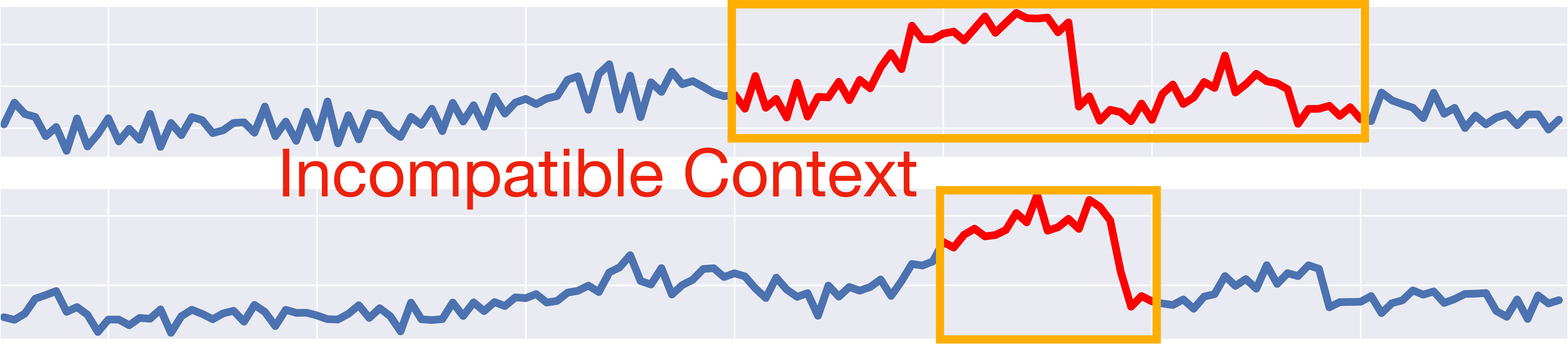}
    \vspace{-12.5pt}
    \caption{Comparison of two machines in SMD dataset.}
    \label{fig:smd}
\vspace{-17.5pt}
\end{figure}

To facilitate knowledge transfer between time series from two different domains, it is critical to align the context of time points (i.e., length of subsequences). Figure~\ref{fig:smd} compares an attribute of two machines from a server monitoring dataset (i.e., SMD dataset~\cite{su2019robust}). We can observe that even if the anomalies of two machines behave similarly, directly conducting knowledge transfer without aligning the context (i.e., the orange window) may lead to negative transfer. To this end, we hypothesize that aligning the context of two domains with different context window sizes benefits anomaly detection. 

\begin{figure}
    \centering
    \includegraphics[width=0.90\linewidth]{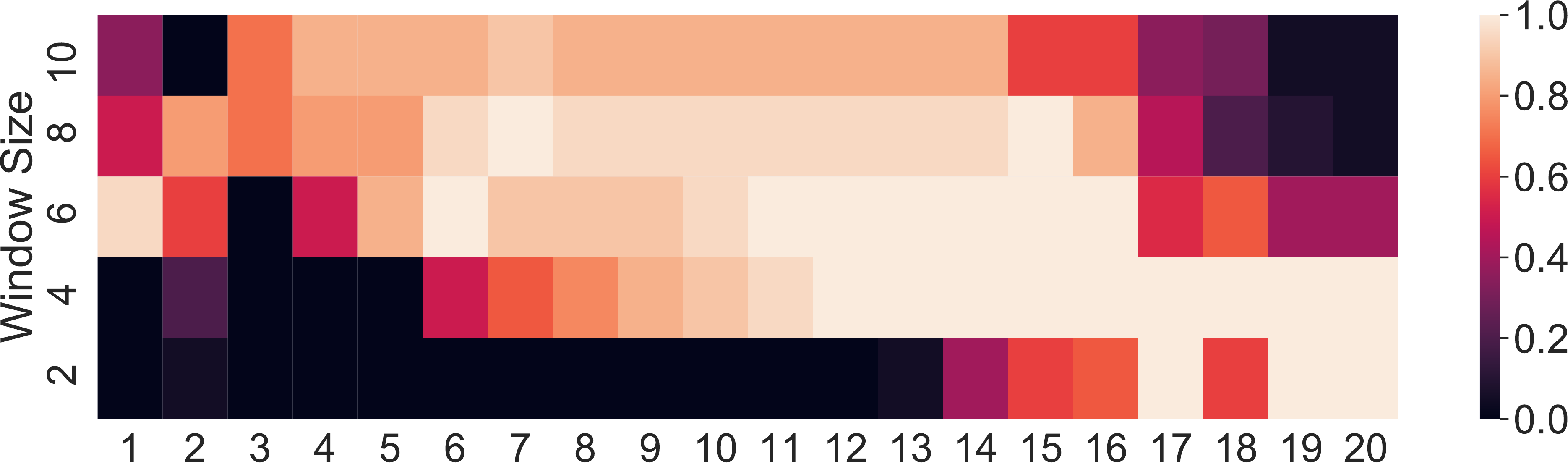}
    \vspace{-10pt}
    \caption{Preliminary context window alignment experiment. The X-axis is the anomaly ID, and the Y-axis is context window size of the target domain. Lighter color indicates better detection ratios.}
    \label{fig:prelim}
\vspace{-12.5pt}
\end{figure}

To validate the assumption, we conduct knowledge transfer between two deep anomaly detectors from two domains and apply multiple context window sizes to the target domain while fixing the source domain window size. Figure~\ref{fig:prelim} illustrates the preliminary experiment that reports the detected ratio of 20 outliers with different context alignment settings for 20 runs. We observe that setting target domain windows at $6$ and $8$ leads to generally better performance than making them the same as the source domain (i.e., window size of $10$). Furthermore, different anomalies can be better detected with different context window sizes. For example, anomalies $1$ and $6$ can always be detected when the window size is $6$, whereas anomalies $2$ and $7$ can be better detected when the window size is set to $8$. The observations above motivate us to develop a contextual domain adaptation framework to facilitate knowledge transfer for time series anomaly detection.

However, it is non-trivial to develop the framework due to two challenges. First, aligning two domains for knowledge transfer while simultaneously modeling the correlation between sensors and temporal dependency between time points is a complex task. As each machine in the SMD dataset has 38 sensors, transferring knowledge between two machines requires extracting domain invariant features from the correlation between 38 sensors from two domains. Second, anomalies from different domains behave differently. It is challenging to extract beneficial information for target domain anomaly detection. For instance, directly exploiting the information of anomalies with extreme values to detect anomalies with sudden drop behavior in a fully supervised manner may lead to sub-optimal performance.


To address the challenges, we propose ContexTDA, a data-centric framework~\cite{zha2023data-centric-perspectives,zha2023data-centric-survey} to select context window size toward time series domain adaptation for anomaly detection. Motivated by recent success of deep reinforcement learning on partial domain adaptation~\cite{chen2020domain,chen2020selective}, we formulate the time series context sampling problem of domain adaptation into a Markov decision process (MDP). A MDP solver is adopted to simultaneously capture temporal dependencies and domain correlations for aligning the context of two domains. Specifically, as the recent benchmark~\cite{schmidl2022anomaly} shows the promising of LSTM autoencoder~\cite{kieu2019outlier}, we instantiate the ContexTDA with the LSTM autoencoder as the base anomaly detector and adopt Deep Q-learning~\cite{mnih2015human} as the MDP solver, to address discrete action space. Extensive experiments suggest the superiority of ContexTDA in transferring knowledge between homogeneous domains, and a pilot study on two heterogeneous datasets demonstrates the potential of ContexTDA toward heterogeneous time series domain adaptation. To sum up, we make the following contributions:
\begin{itemize}[leftmargin=10pt]
    \setlength\itemsep{0.05em}
    \item Identify context sampling as a key toward time series domain adaptation for detecting anomalies.
    
    \item Formulate the context sampling into a MDP and propose ContexTDA to facilitate domain adaptation.
    
    \item Instantiate ContexTDA with DQN and LSTM autoencoder with extensive experiments to show the promise of ContexTDA compared to the state-of-the-arts domain adaptation methods.
\end{itemize}


\section{Preliminary}
In this section, we define the problem of context sampling for time domain adaptation and introduce the background of time series anomaly detection, time series domain adaptation, and deep reinforcement learning.
\vspace{-10pt}
\subsection{Problem Statement}

Let $X=(v_0, v_1 \cdots, v_t)$ and $\hat{X}=(\hat{v}_0, \hat{v}_1 \cdots, \hat{v}_t)$ be two fully observed multivariate time series data of source and target domain respectively. Let $Y \in \mathbb{N}^{t+1}$ be the label of source domain $X$ indicating whether individual time point $v_t$ is an anomaly or not. Time series anomaly detection aims at recognizing if a time point $\hat{v}_t$ is anomalous or not. A scoring function $\mathcal{F}: \hat{x}_t \to \mathbb{R}$ evaluates the degree of anomalous of individual instances based on the input data $\hat{x}_t$; the higher the score, the more anomalous the time point $t$ is. A context window is used to sample subsequences from $\hat{X}$, resulting in the input data $\hat{x}_t=(\hat{v}_{t-m}, \hat{v}_{t-m+1},\cdots, \hat{v}_t)$ as the input of $\mathcal{F}$, where $m$ is the size of the context window. Following the context window mechanism, our goal is to transfer the source domain information (i.e., $X$ and $Y$) with a context sampling policy in order to optimize the performance of $\mathcal{F}$ in detecting anomalies from the unlabeled $\hat{X}$.


The context sampling problem can be formulated as follows. Given two multivariate time series data, our goal is to jointly learn a sampling policy $\widetilde{\pi}$ with an anomaly detector, where $\widetilde{\pi}$ maps $x_t$ and $\hat{x}_t$ into two context window sizes $m$ and $n$. This way, by sampling subsequences $x_t=(v_{t-m}, v_{t-m+1}, \cdots v_{t})$ and $\hat{x}_t=(\hat{v}_{t-n}, \hat{v}_{t-n+1}, \cdots \hat{v}_{t})$ to perform domain adaptation, the performance of anomaly detection on the target domain $\hat{X}$ can be optimized.

\vspace{-10pt}
\subsection{Deep Time Series Anomaly Detection}
To detect anomalies from a time series, existing deep learning approaches assume that normal data instances are compact in hyperspace~\cite{steinwart2005classification}, and therefore model majority patterns lie in the data with autoencoder~\cite{baldi2012autoencoders} to identify the decision boundary between anomalies and normalities. Specifically, a fix-sized sliding window is adopted to segment the time series into subsequences that reflects local patterns of the data. Then, autoencoder~\cite{sakurada2014anomaly} can be introduced to model the majority by minimizing the dissimilarity between the input subsequences and the decoded subsequences based on the low-dimensional latent vector generated by the encoder with following loss function:
\begin{equation}
    \label{eq:recon}
    \min_{\mathcal{D}, \mathcal{E}} ||X - \mathcal{D}(\mathcal{E}(X))||_2^{2},
\end{equation}
where $X \in \mathbb{R}^{t \times w \times n}$ is a tensor representing a segmented $n$-dimensional time series data with window size $w$ and timesteps $t$, $\mathcal{E}$ and $\mathcal{D}$ are the encoder and decoder. This way, the loss value of the potential anomalous subsequence in timestamp $t$ may explicitly larger than rest of the subsequences. Additionally, instead of using multi-layer perceptron, LSTM~\cite{kieu2019outlier} can be adopted to further capture the temporal correlations between individual subsequences. As recent benchmark~\cite{schmidl2022anomaly} shows promise of the LSTM autoendoer, we employ LSTM autoencoder as the core anomaly detector. One may also adopt other RNN-based anomaly detectors~\cite{hundman2018detecting,su2019robust} into the framework.

\vspace{-10pt}
\subsection{Time Series Domain Adaptation}
Domain adaptation is widely adopted to image data due to the consistent definition of each entry in images. The two common strategies to perform domain adaptation are domain discrepancy minimization~\cite{long2013transfer} and domain discrimination~\cite{tzeng2017adversarial,long2015learning}. Domain discrepancy minimization introduces a mapping function $\mathcal{M}$ to map source and target domain data into a shared subspace, and then perform distance minimization based on the feature vectors of both domains in the subspace:
\vspace{-5pt}
\begin{equation}
    \label{eq:mmd}
     \min_{\mathcal{M}} ||\mathcal{M}(X) - \mathcal{M}(\hat{X}))||_{2}^{2}
     \vspace{-10pt}
\end{equation}

where $X$ and $\hat{X}$ are the source and target domain data. Recent study~\cite{cai2021time} on domain adaptation for time series fault detection develop a LSTM-based classification framework that generates source and target domain features by a shared LSTM unit and adopt domain discrepancy minimization with self-attention mechanism to adaptively align the size of subsequences between two domains toward domain adaptation.

Domain discrimination conducts adversarial learning with generator and discriminator. The goal of the generator $\mathcal{G}$ is to generate domain invariant features that cannot be distinguished by the discriminator $\mathcal{H}$ while the target of discriminator is to identify the domain of the input feature vector from the generator: 
\vspace{-7.5pt}
\begin{equation}
    \small
    \label{eq:dd} \ \ \
    \min_{\mathcal{H}} \max_{\mathcal{G}} \mathbb{E}_{x \sim X\cup \hat{X}}[\mathcal{H}(x)] + \mathbb{E}_{z \sim Z}[\log(1-\mathcal{H}(\mathcal{G}(z)))]
    \vspace{-10pt}
\end{equation}
where $z \in Z$ is a prior noise. By performing the min-max optimization between $\mathcal{G}$ and $\mathcal{H}$, the generator will be able to generate domain invariant features. Latest study~\cite{jin2021attention} on domain adaptation for time series forecasting introduces a dual encoder-decoder framework with a shared attention layer as the feature generator to generate domain invariant features to train a domain discriminator. However, existing works are tailored for classification and forecasting tasks, which focus on transferring the knowledge of majority distribution of time series and potentially ignore the diverse minority distribution~\cite{zha2022towards2}. Therefore, they cannot be directly adopted to anomaly detection problem. Motivated by partial domain adaptation, in our framework, we introduce a context sampling policy to tailor the contextual information for individual time points to alleviate the aforementioned issues. 
\vspace{-10pt}
\subsection{Solving Markov Decision Process via Deep Reinforcement Learning}
\label{sec:mdp}
Markov Decision Process~(MDP) models sequential decision making process by a quintuple $(\mathcal{S}, \mathcal{A}, \mathcal{P}_{T}, \mathcal{R}, \gamma)$, where $\mathcal{S}$ is a finite set of states, $\mathcal{A}$ is a finite set of actions, $\mathcal{P}_{T}: \mathcal{S} \times \mathcal{A} \times \mathcal{S} \to\mathbb{R}^+$ is the state transition probability function 
that maps the current state $s$, action $a$ and the next state $s'$ to a probability value, $\mathcal{R}: \mathcal{S} \to \mathbb{R}$ is the immediate reward function that reflects the quality of action $a$, and $\gamma \in (0,1)$ is a discount factor. At each timestep $t$, the agent takes action $a_t \in \mathcal{A}$ based on the current state $s_t \in \mathcal{S}$, and observes the next state $s_{t+1}$ as well as a reward signal $r_t = \mathcal{R}(s_{t+1})$. The goal is to search an optimal series of actions such that the expected discounted cumulative reward is maximized. Mathematically speaking, we would like find a policy $\pi: \mathcal{S} \to \mathcal{A}$ to maximize $\mathbb{E}_\pi[\sum_{t=0}^{\infty}\gamma^t r_t]$.

Deep reinforcement learning algorithms are designed to solve the Markov decision process (MDP) with deep neural networks. In this work, we adopt model-free deep reinforcement learning, which learns the decision function during the exploration. Deep-Q Learning (DQN)~\cite{mnih2015human} is a pioneering algorithm which uses deep neural networks as a function approximator to model state-action values Q(s, a) that satisfies:
\vspace{-5pt}
\begin{equation}
     \label{eq:qlearning}
      Q(s,a) = \mathbb{E}_{s'} [\mathcal{R}(s') + \gamma \max_{a'} (Q(s', a')],
     \vspace{-10pt}
\end{equation}
where $s'$ is the next state and $a'$ is the next action. DQN introduces two techniques to stabilize the training: (1) a replay buffer to reuse past experiences; (2) a separate target network that is periodically updated. In this work, we employ DQN to solve the MDP; one could also apply advanced algorithms such as soft actor-critic~\cite{haarnoja2018soft}.



\begin{figure*}
    \centering
    \includegraphics[width=0.8\linewidth]{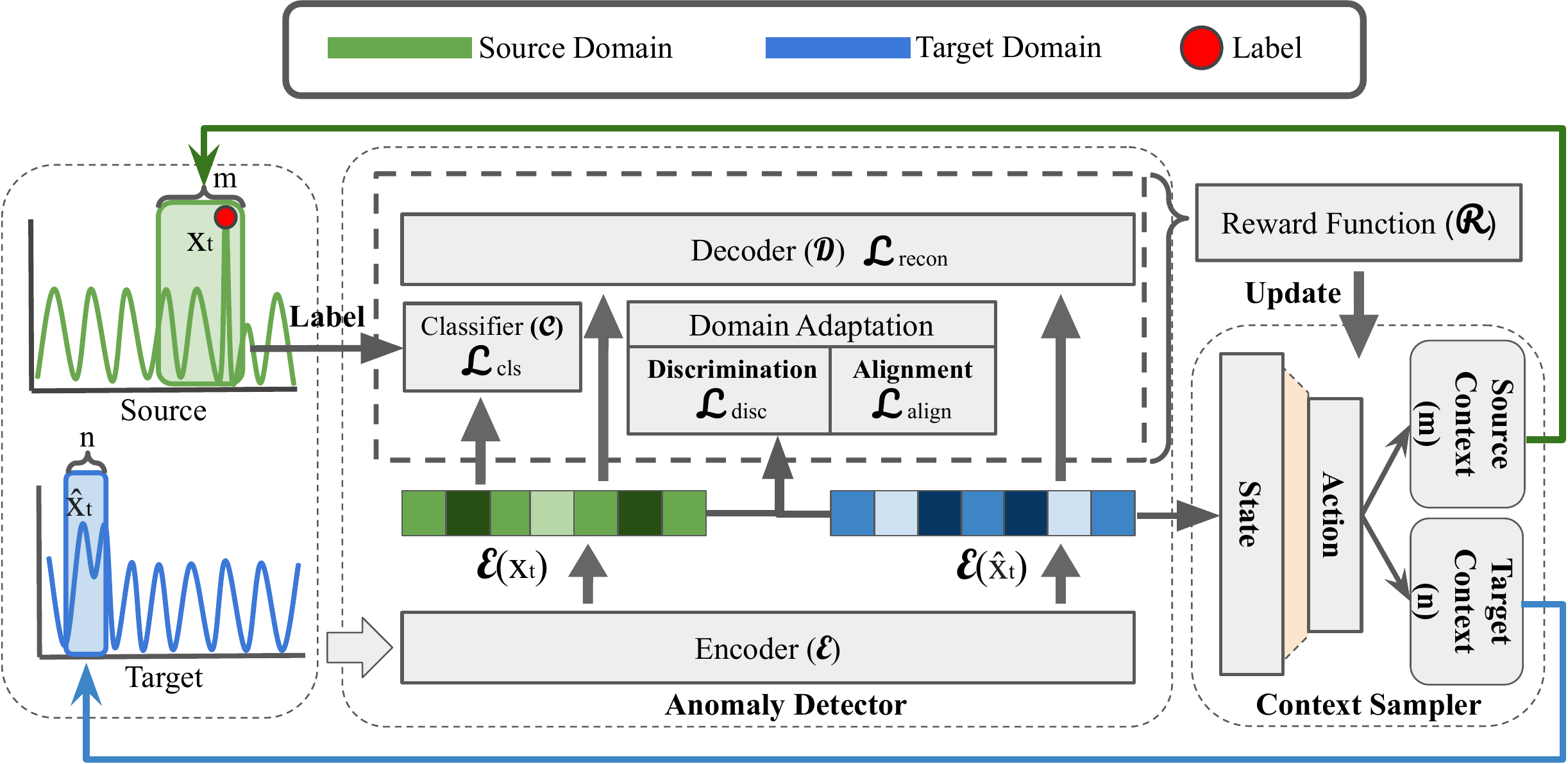}
    \vspace{-10pt}
    \caption{An illustration of ContextTDA. In each step, sub-sequences from two domains are encoded as feature vectors. The feature vectors are adopted to compute losses for updating the anomaly detector and source domain classifier, and to compute the reward signals. Then, the context sampler generates context window sizes of two domains for the next time step based on the feature vectors and is updated with the reward signal.}
    \label{fig:overview}
    \vspace{-15pt}
\end{figure*}

\vspace{-7.5pt}
\section{Methodology}
Figure~\ref{fig:overview} illustrates the overview of the proposed framework. There are two main components: a context sampler and an anomaly detector. The context sampler aims at learning an optimal policy to sample contextual information for individual time points as input data batches for the anomaly detector. The anomaly detector takes the output of the context sampler to model the data from two domains into two feature vectors for knowledge transfer and anomaly detection.

To integrate the two modules, we frame the time series domain adaptation into a Markov decision process. Specifically, the context sampler considers the feature vectors generated by the anomaly detector as the state. It maps the state into the action that specifies the context window sizes for the source and target domains in the next timestamp. Then, the input data of the anomaly detector will be sampled according to the action of the context sampler. During the training process, three types of losses will be computed to generate reward signals for updating the context sampler while training the anomaly detector: classification loss of source domain data to better exploit label information; unsupervised reconstruction loss of source and target domains to reflect majority modeling for anomaly detection task; and domain adaptation losses to encourage knowledge transfer between domains.

In what follows, we elaborate on the details of our framework. We first focus on framing the context sampling problem into a Markov decision process. Then, we introduce the domain adaptation strategies and model training details. Lastly, we introduce anomaly inference with the trained model.

\vspace{-10pt}
\subsection{Context Sampling as Markov Decision Process}
Viewing each single data point alone neglects temporal dependency~\cite{zolhavarieh2014review} and thus a context window is usually adopted to capture meaningful local patterns. In this session, we define time series context sampling as a Markov decision process to capture local patterns to facilitate domain adaptation. As discussed in Section~\ref{sec:mdp}, the key components of an MDP include states, actions, and rewards, as well as a transition function that maps the current state and action into the next state. Here, we define our state, action, and reward signal as follows:
\vspace{-5pt}
\begin{itemize}[leftmargin=10pt]
    \vspace{-7.5pt}
    \item \textbf{State:} The state $s_t \in \mathcal{S}$ at timestep $t$ is defined as a $2n$ dimensional vector $(\mathcal{E}(x_t), \mathcal{E}(\hat{x}_t))$, which is the concatenation of two $n$-dimensional feature vectors from the encoder $\mathcal{E}$ of the anomaly detector based on current data point of source $x_t$ and target $\hat{x}$.
    \vspace{-7.5pt}
    \item \textbf{Action:} The action $a_t \in \mathcal{A}$ at timestep $t$ is a two-dimensional vector. The first and second dimensions are the window size for source and target domains respectively. The action space for the source and target domains ranges from 1 to a given maximum window size.
    \vspace{-7.5pt}
    \item \textbf{Reward:} We define the reward $r_t$ at timestep $t$ as a combinatorial signal of source domain classification loss, source and target domain reconstruction loss, domain alignment and discrimination loss. The goal is to encourage the context sampler to sample quality data batches toward better exploitation of label information and domain adaptation.
\end{itemize}
\vspace{-5pt}
The proposed domain adaptation process consists of three phases: 1) generate state $s_t$ based on the source and target into windows that ends at timestamp $t$, 2) generate an action $a_t$ from $\widetilde{\pi}(s_t)$ to specify the context window size of the next data points from source and target domains, 3) sample the next input data for training the anomaly detector and the source classifier based on $a_t$ and generate the next state $s_{t+1}$. 

To solve the aforementioned MDP, we employ deep reinforcement learning to model the transition dynamics between states and actions and learn a context sampling policy for domain adaptation. Due to the fact that window sizes for the two domains are finite integers, the action space is a discrete space and actions for the two domains are always positive integers. Therefore, we introduce deep Q-learning to address the problem. In particular, the reward function is the critical signal that guides the whole learning process. We design the reward function as the reciprocal of the combination of multiple losses, which will be elaborated in Section~\ref{sec:reward}. With the reward function, the Q-function are approximated by a multi-layer perceptron, and the epsilon-greedy algorithm is adopted to form the sampling policy $\widetilde{\pi}$.

\vspace{-10pt}
\subsection{Contextual Domain Adaptation}
\label{sec:reward}
To facilitate knowledge transfer from a well-labeled source domain to an unlabeled target domain, we design a combinatorial reward function that reflects the status of the anomaly detector, the exploitation of the source domain label information, and the discrepancy between the two domains. The goal of the reward function is to encourage the context sampler to obtain an adequate context window size such that the label information can be properly transferred to the target domain and domain discrepancies can be minimized. Specifically, we involve three types of losses generated by the LSTM anomaly detector in the reward function: 1) classification loss on the source domain; 2) reconstruction losses on both domains; and 3) domain adaptation losses between two domains. In the meantime, the encoder and decoder of the LSTM anomaly detector are jointly trained with a source domain classifier and a domain discriminator.

First, to compute the classification loss, we introduce a multi-layer perceptron classifier $\mathcal{C}$ with sigmoid activation function, and the input of latent features generated by the encoder of the anomaly detector. Then, we compute the binary cross-entropy on the source domain and re-weight the loss values with weight matrix $w_t \in W$ on labeled anomalies and nomalities to form the reward signal that puts emphasis on precise classification on anomalies:
\vspace{-5pt}
\begin{displaymath}
    \resizebox{.99\hsize}{!}{$\mathcal{L}_{cls} = \sum_{x_t \in X} -w_t [y_t \cdot \log(\mathcal{C}(\mathcal{E}(x_t))) + (1-y_t) \cdot \log(1-\mathcal{C}(\mathcal{E}(x_t)))]$}.
    \vspace{-5pt}
\end{displaymath}
where weight matrix $w_t \in W$ specifies the weight for anomalies and normalities.

Second, we compute the reconstruction loss to encourage the detector to precisely model the general distribution of both domains. To leverage the label information in the source domain, we follow the equation~\ref{eq:recon} to compute reconstruction loss in both domains:
\vspace{-5pt}
\begin{displaymath}
    \small
    \mathcal{L}_{recon} =  \sum_{x_t \in X^{+}, \hat{x}_t \in \hat{X}} ||x_t - \mathcal{D}(\mathcal{E}(x_t))||_{2}^{2} + ||\hat{x}_t - \mathcal{D}(\mathcal{E}(\hat{x}_t))||_{2}^{2}
    \vspace{-5pt}
\end{displaymath}
where $X^{+}$  and $\hat{X}$ are the labeled normalies on source domain and the unlabeled target domain dataset. This way, we can put emphasis on the data distribution of the source domain without outliers, which is potentially beneficial to modeling the general distribution of the target domain.

Finally, to facilitate domain adaptation, we compute domain discrepancy minimization loss and domain discrimination loss to encourage the encoder to generate domain-invariant features toward minimum domain discrepancy. We follow the equation~\ref{eq:mmd} to compute the domain discrepancy with the feature vectors generated by the encoder of the anomaly detector:
\vspace{-5pt}
\begin{displaymath}
     \mathcal{L}_{align} = \sum_{x_t \in X, \hat{x_t} \in \hat{X}}||\mathcal{E}(x_t) - \mathcal{E}(\hat{x_t}))||_{2}^{2}.
     \vspace{-7.5pt}
\end{displaymath}
Following the equation~\ref{eq:dd}, we introduce a domain classifier $\mathcal{K}$ with sigmoid activation function to compute the domain discrimination loss based on the feature vectors generated by the encoder:
\vspace{-5pt}
\begin{displaymath}
   \resizebox{.99\hsize}{!}{$\mathcal{L}_{disc} = -\sum\limits_{x_t \in X \cup \hat{X}} \tilde{y}_t \cdot \log(\mathcal{K}(\mathcal{E}(x_t))) + (1-\tilde{y}_t) \cdot \log(1-\mathcal{K}(\mathcal{E}(x_t)))$}
   \vspace{-5pt}
\end{displaymath}
where $\tilde{y}_t \in \tilde{Y}$ is the label indicating the domain of individual data instances $x \in X \cup \hat{X}$.

Here, since the domain discrimination loss will be minimized to prompt the encoder to generate domain-distinctive representations, we set the $L_{disc}$ with a negative coefficient to encourage the context sampler to sample subsequences that maximize the domain discrimination loss. This way, when DQN maximizes the reward signal, the loss will also be maximized, which leads to min-max optimization that generates domain invariant feature vectors. Based on the losses, the reward function is defined as:
\vspace{-5pt}
\begin{equation}
     \resizebox{.925\hsize}{!}{\ \ \ \ \ $\mathcal{R} (s_t, a_t) = \frac{1}{\alpha \cdot L_{cls} + \beta \cdot L_{recon} + \gamma \cdot L_{align} - \lambda \cdot L_{disc}}$},
    \label{eq:reward}
    \vspace{-2.5pt}
\end{equation}
where $\alpha$, $\beta$, $\gamma$, $\lambda$ are the hyper-parameters to control the reward. The details of the algorithm and training procedure are provided in Appendix~\ref{appendix:algorithm}.


\vspace{-10pt}
\subsection{Anomaly Inference}
To detect the anomalies in target domain, we integrate the trained anomaly detector with the learned sampling policy $\widetilde{\pi}$. Specifically, we sample the context window sizes for individual time points with the learned policy $\widetilde{\pi}$, and input the sampled context window to the anomaly detector to detecting anomalies. During the training procedure, we introduce an action counter $\mathcal{N}$ to identify the most frequently selected context window size on source domain and correspondingly sample the data to generate $\mathcal{E}(x_t)$ for all time points in source domain. Then, the $\mathcal{E}(x_t)$ will be concatenated with the $\mathcal{E}(\hat{x}_t)$, which generated by the encoder based on the context window sampled with policy $\widetilde{\pi}$ for target domain. 

We iteratively sample $x_t$ and $\hat{x_t}$ based on action $a_t$ to obtain the state $s_t$ by fixing the source domain action as the most frequently selected context window size to evaluate the level of anomaly for each time point. The intuition behind this is to prompt the context sampler to focus on the context sampling for target domain data. The feature vector of the target domain $\mathcal{E}(\hat{x_t})$ is fed to the source classifier $\mathcal{C}$ during each iteration to obtain the confidence score of being an anomaly. Meanwhile, the reconstruction error is computed based on equation~\ref{eq:recon} to measure the level of deviation from the general distribution of the data. Finally, the anomalous score can be computed by multiplying the reconstruction error with the confidence score:
\vspace{-5pt}
\begin{displaymath}
    A(\hat{x}_t) = \mathcal{C}(\mathcal{E}(\hat{x}_t)) \cdot ||\hat{x}_t - \mathcal{D}(\mathcal{E}(\hat{x}_t))||_2^{2}
\end{displaymath}


\vspace{-15pt}
\section{Experiment}
We conduct experiments to answer the following research questions:
\textbf{RQ1}: How does the ContexTDA compare against existing time series anomaly detection algorithms and the state-of-the-art time series domain adaptation methods when two domains share identical feature space? 
\textbf{RQ2}: How does the ContexTDA compare against baselines when two domains are equipped with two entirely different feature spaces? 
\textbf{RQ3}: How do hyper-parameters affect the ContexTDA and what are the intuitions for tuning hyper-parameters?

\vspace{-10pt}
\subsection{Experiment Settings}
\textbf{Datasets.} Domain adaptation can be homogeneous or heterogeneous. The knowledge from a source domain with identical data characteristics (i.e., data dimensions) is transferred to the target domain in the homogeneous setting, whereas the knowledge from the source domain with completely different characteristics from the target domain is adapted in the heterogeneous setting. We adopt \textbf{Server Machine Dataset (SMD)}~\cite{su2019robust} and \textbf{Boiler}~\cite{cai2021time} for homogeneous setting; and \textbf{Mars Science Laboratory rover (MSL)} and \textbf{Soil Moisture Active Passive satellite (SMAP)}~\cite{hundman2018detecting} for the heterogeneous setting. More details are provided in Appendix~\ref{appendix:data}.

In the homogeneous experiments, we adopt the first set of machines in the SMD dataset and select the first machine as the source domain and use the rest as the target domains. For the Boiler dataset, we use all 3 boilers and perform domain adaptation on all pair-wise combinations. In the heterogeneous setting, due to the data sparsity of MSL, we use the SMAP as the source domain and MSL as the target domain.



\begin{table*}[]
\scriptsize
\centering
\begin{tabular}{@{}c|c|cc|ccc|cc@{}}
\toprule
 \multicolumn{2}{c|}{Algorithm Type}  & \multicolumn{2}{c|}{Single Domain}  &  \multicolumn{5}{c}{Dual Domain}   \\ \cmidrule{1-9}
 Dataset & Macro-F1 / AUC & AE-MLP & AE-LSTM & RDC & RDC-VRADA & SASA & RandContexTDA & ContexTDA \\\midrule
 & 1 $\rightarrow$ 2   & 0.72 / 0.83 & 0.74 / 0.90 & 0.74 / 0.89 & 0.74 / 0.91 & 0.59 / 0.63 & 0.61 / 0.87 & \textbf{0.75} / \textbf{0.91} \\
& 1 $\rightarrow$ 3   & 0.57 / 0.70 & 0.49 / 0.41 & 0.57 / 0.72 & 0.49 / 0.41 & \textbf{0.61} / \textbf{0.90} & 0.54 / 0.71 & 0.57 / 0.75 \\
& 1 $\rightarrow$ 4   & 0.55 / 0.74 & 0.54 / 0.41 & 0.54 / 0.75 & 0.54 / 0.41 & 0.55 / 0.75 & 0.52 / 0.72 & \textbf{0.59} / \textbf{0.76} \\
& 1 $\rightarrow$ 5   & 0.54 / 0.79 & 0.55 / 0.84 & 0.56 / 0.85 & 0.55 / 0.79 & 0.65 / 0.87 & 0.54 / 0.81 & \textbf{0.66} / \textbf{0.87} \\
SMD & 1 $\rightarrow$ 6   & 0.71 / 0.88 & 0.71 / 0.91 & 0.71 / 0.87 & 0.71 / \textbf{0.91} & 0.44 / 0.84 & 0.71 / 0.81 & \textbf{0.73} / 0.84 \\
& 1 $\rightarrow$ 7   & 0.48 / 0.55 & 0.48 / 0.50 & 0.49 / 0.54 & 0.48 / 0.50 & 0.31 / \textbf{0.57} & 0.46 / 0.51 & \textbf{0.51} / 0.53 \\
& 1 $\rightarrow$ 8   & 0.55 / 0.57 & 0.53 / \textbf{0.70}  & 0.55 / 0.58 & 0.54 / 0.56 & 0.52 / 0.56 & 0.54 / 0.43 & \textbf{0.58} / 0.58 \\ 
\cmidrule{2-9}
& Avg. & 0.59 / 0.72 & 0.58 / 0.67 & 0.59 / 0.74 & 0.58 / 0.64 & 0.52 / 0.73 & 0.56 / 0.69 & \textbf{0.63} / \textbf{0.75} \\
\bottomrule\toprule
& 1 $\rightarrow$ 2  & 0.44 / 0.64 & 0.43 / 0.48 & 0.43 / 0.54  & 0.43 / 0.48  & \textbf{0.53} / \textbf{0.88} & 0.42 / 0.48 & 0.50 / 0.59 \\
& 1 $\rightarrow$ 3  & 0.40 / 0.28 & 0.40 / 0.11 & 0.43 / 0.49 & 0.42 / 0.18 & 0.41 / 0.48 & 0.31 / 0.36 & \textbf{0.50} / \textbf{0.67} \\
 & 2 $\rightarrow$ 1  & 0.39 / 0.18 & 0.40 / 0.21 & 0.40 / 0.36 & 0.40 / 0.15  & \textbf{0.53} / \textbf{0.90} & 0.44 / 0.60 & 0.51 / 0.66 \\
Boiler & 2 $\rightarrow$ 3  & 0.40 / 0.38 & 0.40 / 0.20 & 0.45 / 0.39 & 0.42 / 0.21 & 0.41 / 0.49 & 0.33 / 0.36 & \textbf{0.50} / \textbf{0.69} \\
& 3 $\rightarrow$ 1  & 0.39 / 0.20 & 0.40 / 0.16 & 0.39 / 0.31 & 0.40 / 0.15 & 0.48 / 0.67 & 0.42 / 0.65 & \textbf{0.51} / \textbf{0.67} \\
& 3 $\rightarrow$ 2  & 0.48 / 0.54 & 0.49 / 0.48 & 0.49 / 0.55  & 0.49 / 0.48 & 0.46 / 0.31 & 0.48 / 0.48 & \textbf{0.50} / \textbf{0.57} \\\cmidrule{2-9}
 & Avg. & 0.42 / 0.37 & 0.42 / 0.27 & 0.43 / 0.44 & 0.43 / 0.28 & 0.47 / 0.62 & 0.40 / 0.49  & \textbf{0.50} / \textbf{0.64} \\
\bottomrule
\end{tabular}
\vspace{-14pt}
\caption{Homogenous adaptation on SMD and Boiler dataset.}
\label{tab:homo}
\vspace{-15pt}
\end{table*}

\textbf{Baselines.} We consider two competitive autoencoders suggested by a recent benchmark\cite{schmidl2022anomaly} with different neural units (i.e., \textbf{AE-MLP}~\cite{sakurada2014anomaly}, \textbf{AE-LSTM)~\cite{malhotra2016lstm}} to examine the anomaly detection performance in the target domain. We compare ContexTDA with state-of-the-art domain adaptation methods (i.e., \textbf{RDC}~\cite{tzeng2014deep}, \textbf{RDC-VRADA}~\cite{purushotham2016variational}, \textbf{SASA}~\cite{cai2021time}) to evaluate the effectiveness of knowledge transfer in our framework. We also create a simplified version of our framework with a random context sampling policy (i.e., \textbf{RandContexTDA}) to examine the effectiveness of policy learning. Details of baselines can be found in Appendix~\ref{appendix:baseline}.

\textbf{Evaluation Protocol.} Following the imbalanced classification~\cite{johnson2019survey,he2009learning}, we evaluate the performance via the macro-averaged F1-score (Macro-F1) and the area under the ROC Curve (AUC).

\vspace{-10pt}
\subsection{Homogeneous Transfer} 
To study \textbf{RQ1} we compare ContexTDA with baselines on the performance of time series anomaly detection. Table~\ref{tab:homo} reports the Macro-F1 and AUC of all algorithms across all pairs of settings on the two datasets. In general, ContexTDA outperforms the second best baseline with 6.7\% and 6.3\% improvement on the average Macro-F1 score and achieves a superior average AUC score to all baselines. Additionally, ContexTDA achieved superior performance on 6 out of 7 settings on the SMD dataset and 4 out of 6 settings on the Boiler dataset. We make the following observations:

First, transferring the knowledge with a unified context window size for both domains may not be helpful for anomaly detection. Specifically, we observe that the dual-domain baselines are generally outperforming AE-LSTM on AUC score. However, by comparing RDC, RDC-VRADA, and SASA with AE-LSTM on the Macro-F1 score, the performances are inferior or comparable to the AE-LSTM. The phenomenon suggests that although transferring the knowledge without tailored contextual information for each point is able to improve the quality of the decision score, the improvements are mainly on normalities instead of anomalies.

Second, aligning the context size locally for domain adaptation facilitates knowledge transfer on both outliers and normalities. When comparing the proposed ContexTDA with dual domain baselines, we observe that it slightly improves the AUC score and achieves the superior Macro-F1 score among all baselines. In particular, when comparing ContexTDA with SASA, we observe that the AUC score is generally comparable while Macro-F1 is significantly better. This suggests that the ContexTDA framework is capable of extracting meaningful representation for each time point such that both normal and anomalous information can be properly transferred.


Third, combining MMD with domain discrimination requires careful context alignment. By comparing RDC with RDC-VRADA, we observe that RDC-VRADA is generally inferior to RDC, which demonstrates that directly combining two domain adaptation methods without proper context alignment leads to negative transfer on both normalities and anomalies. In addition, we observe that ContexTDA, which is built upon RDC-VRADA, gains significant improvement over RDC-VRADA. This suggests that personalized contextual information for each point is required to combine the two domain adaptation methods. The reason behind it may be due to the different behavior of anomalies. Extracting domain-invariant features with identical contexts for anomalies with entirely different behavior from two different domains may lead to noisy low-dimensional representation and therefore lead to negative transfer.

Lastly, the proposed context sampling policy can be properly learned by the DQN. Specifically, we can observe a significant performance improvement across all datasets and settings when comparing ContexTDA with RandContexTDA. This suggests that the proposed Markov decision process is an optimizable objective and the reward function is capable of guiding the policy learning process toward performance improvement with domain adaptation, and therefore, the adopted DQN indeed learns an effective context sampling policy.

We notice that ContexTDA cannot outperform SASA in a few cases. We conduct an investigation and observe that the behaviors of anomalies across two domains are highly consistent (i.e., anomalies are triggered by similar sets of attributes with highly similar values). This suggests supervised learning would be favored when anomalous behaviors from two domains are highly similar. The reason behind this is that highly similar behaviors of anomalies lead to a concentrated distribution of anomalies and therefore form two clusters in feature space, normalies and anomalies, that are favorable to binary classification settings.

\vspace{-10pt}
\subsection{Heterogeneous Transfer}
To answer \textbf{RQ2}, we compare ContexTDA with both single and dual domain baselines on SMAP and MSL datasets. Since SASA is designed for the setting of sharing feature space between domains, it is not applicable to this setting. For dual domain baselines, we implement two encoder-decoder networks for modeling the two domains respectively and exploit the representations generated by the two encoders to perform domain alignment and domain discrimination. In general, ContexTDA outperforms the second-best dual domain baseline with a $9.9\%$ improvement on Macro-F1 and a $1.6\%$ on AUC score. Based on the Table~\ref{tab:hetero} we make the following observations:

First, domain discrimination facilitates heterogeneous domain adaptation. Comparing RDC with AE-LSTM, we can observe that forcefully aligning two entirely different domains leads to significant negative transfer. However, by comparing RDC-VRADA with AE-LSTM and RDC, we can learn that adding domain discrimination to extract domain invariant features alleviates the negative transfer.

Second, locally sampled context information for individual time points allows the model to focus on beneficial information. Comparing ContexTDA to all the domain adaptation frameworks, it is the only one without a negative transfer effect on AUC while gaining improvement on Macro-F1 from AE-LSTM. We observe a phenomenon that the variance of selected window sizes is higher than the variances in homogeneous settings. It is likely that the context sampler reduces context information that hinders knowledge transfer while amplifying the information that benefits knowledge transfer in order to achieve better performance. 

\vspace{-7.5pt}
\begin{table}[!h]
\scriptsize
\centering
\begin{tabular}{c|c|cc}
\toprule
 \multicolumn{2}{c|}{Algorithms} & Macro-F1 & AUC \\ \midrule
\multirow{2}{*}{Single Domain} & AE-MLP & 0.53 & 0.48 \\ 
 & AE-LSTM & 0.53 & 0.61 \\ \midrule
\multirow{3}{*}{Dual Domain}& RDC & 0.50 & 0.51 \\ 
& RDC-VRADA & 0.51 & 0.60 \\ 
& ContexTDA & \textbf{0.56} & \textbf{0.61} \\ \bottomrule
\end{tabular}
\vspace{-9pt}
\caption{Heterogeneous adaptation on SMAP$\rightarrow$MSL.}
\label{tab:hetero}
\vspace{-20pt}
\end{table}





\subsection{Ablation Study and Hyperparameter Tuning} 
To study \textbf{RQ3}, we conduct reward ablation study on transferring knowledge from machine 1 to machine 4 of the SMD dataset. Specifically, we remove the individual loss functions that form the reward signal to observe the performance changes in Table~\ref{tab:abla}. We observe that, compared to other ablations, the ablations on the two domain adaptation objectives do not drastically change the performance, which suggests that the majority distribution of two domains is very similar. This phenomenon is reflected in the two datasets that show that the first and fourth machines indeed share a similar set of attributes equipped with constant zero values. Another observation is that the source label information plays a significant role in detecting anomalies rather than reconstruction objective under this setting. One possible explanation is that the anomalies in the two domains share similar behaviors, which is also reflected in the dataset that the anomalies in the two datasets are global anomalies triggered by surges on similar sets of attributes.

Since each objective plays a unique role during the training process and may affect the performance for each setting differently, we discuss the intuitions of hyperparameter tuning for reward signal generation here based on empirical observations. 1.) $\alpha$ should be larger when anomaly behaviors between two domains are similar. 2.) When labeled anomalies are scarce, $\beta$ focuses on general pattern modeling for each domain, and a larger value is required. 3.) When the general patterns of two domains are similar, $\gamma$ improves performance, and a larger value can be used when the two domains are homogeneous. 4.) $\lambda$ facilitates extracting domain invariant features and, therefore, larger value may be useful when two domains are heterogeneous.

\vspace{-5pt}
\begin{table}[!h]
\scriptsize
\centering
\begin{tabular}{c|cc}
\toprule
Ablations & Macro-F1 & AUC \\ \midrule
 w/o source label information ($\alpha=0$) & 0.52 & 0.73 \\
 w/o reconstruction objective ($\beta=0$) & 0.55 & 0.75 \\ 
 w/o domain alignment ($\gamma=0$) & 0.58 & 0.74  \\ 
 w/o domain discrimination ($\lambda=0$) & 0.57  & 0.74 \\ \midrule
 Full ContexTDA & 0.59 & 0.76 \\ \bottomrule
 
\end{tabular}
\vspace{-9pt}
\caption{Ablation study on SMD 1$\rightarrow$4.}
\label{tab:abla}
\vspace{-15pt}
\end{table}
\vspace{-5pt}

\section{Related Work} 
We discuss the connections between the proposed framework and related works on partial domain adaptation. 
\vspace{-16pt}

\subsection{Domain Adaptation}
Domain adaptation~\cite{wilson2020survey,tan2018survey} transfers knowledge learned
from a source domain with label information to the target domain without sufficient labels. Maximum Mean Discrepancy (MMD)~\cite{tzeng2014deep,tzeng2014deep} reduces domain discrepancy within a metric space. Domain discrimination is built upon adversarial learning framework~\cite{goodfellow2014generative,ganin2015unsupervised,tzeng2017adversarial} to generate domain-invariant features that cannot be discerned by a domain discriminator. Adversarial Discriminative Domain Adaptation~\cite{tzeng2017adversarial} proposes an unified framework for adversarial domain adaptation. We integrate MMD with domain discrimination to minimize domain discrepancies with domain-invariant features .
\vspace{-8pt}

\subsection{Time Series Domain Adaptation}
To perform domain adaptation on time series data, existing methods leverage neural architectures such as RNN-LSTM~\cite{purushotham2016variational,chung2015recurrent} and self-attention~\cite{cai2021time,jin2021attention} as feature extractors to perform domain adaptation with MMD, domain discrimination objectives from different level of observations (e.g., time point or sub-sequence), or disentanglement~\cite{li2022towards}. However, all existing approaches are designed for time series regression and classification~\cite{zha2022towards} problems that focus on aligning the majority distribution of two domains and may potentially result in negative transfer on minority distribution. As transferring knowledge of anomalies between two domains requires aligning minority distribution of time series and the anomaly label spaces from two domains often share very limited similarity, all existing methods are unable to be adopted to solve the anomaly detection problem.

\vspace{-8pt}

\subsection{Partial Domain Adaptation}
Partial domain adaptation aims at developing instance-wise selective transfer strategies to alleviate the strong assumption of identical label space for different domains. Re-weighting methods derives various strategies to weigh the source samples according to the class probabilities from domain discriminators~\cite{cao2018partial} or domain-adversarial network~\cite{zhang2018importance}. Other approaches~\cite{chen2020domain,chen2020selective} focus on iterative instance selection methods formulates the instance selection problem into a Markov decision process and adopt deep reinforcement learning algorithms to go through individual instances from different level of observations for partial domain adaptation. However, instance-wise selective transfer methods are not applicable to time series data since there exist temporal dependencies between instances and arbitrary selection of partial instances may fail to consider the dependencies during domain adaptation. Therefore, we propose a contextual domain adaptation framework that focuses on context sampling for individual instances in time series data.

\section{Conclusion}

We propose a data-centric context-aware domain adaptation framework, named ContexTDA, for time series anomaly detection. We formulate the context sampling problem into a Markov decision process and introduce a deep reinforcement learning algorithm with a tailored reward function to facilitate knowledge transfer for time series anomaly detection. The empirical evaluation of the proposed method on two public datasets demonstrates its superiority in homogeneous knowledge transfer. Furthermore, the pilot study on two entirely different time series data reveals that aligning the context window size for two domains may be a key factor toward heterogeneous knowledge transfer.
\bibliographystyle{siamplain}
\bibliography{ref}
\clearpage
\newpage
\appendix

\section{Algorithm Details}
\label{appendix:algorithm}

Algorithm~\ref{alg:overview} illustrates the training details of the proposed framework. The training procedure begins with generating the state of the first timestamp $s_0$ with encoder $\mathcal{E}$ from a pair of randomly sampled context window sizes. The DQN agent will then use a greedy algorithm to sample the corresponding action $a_t$ based on the Q-function in order to determine the context window size for the next timestamp and store the transition in its memory buffer~\cite{zha2019experience} in order to optimize the Q-function. During the training procedure, the performed action will be recorded by an action counter $\mathcal{N}$ for anomaly inference in the target domain.
\setlength{\textfloatsep}{5pt}
\begin{algorithm}[t]
\small
\caption{Training ContexTDA}
\label{alg:overview}
\begin{algorithmic}[1]
\STATE \textbf{Input:} Maximum context window size $K$, number of training epoch $E$, DQN training step $S$, epsilon probability $\epsilon$, reward coefficients $\alpha$, $\beta$, $\gamma$, $\lambda$, number of timesteps $T$.
\STATE Initialize action counter $\mathcal{N}$; $\mathcal{E}$, $\mathcal{D}$, $\mathcal{C}$, $\mathcal{K}$ for anomaly detector; and  Q-function $Q$, memory buffer $\mathcal{B}$ for DQN.
\STATE Randomly sample context widow sizes, generate a state $s_{0}$ by the encoder $\mathcal{E}$.
\FOR {$e$ = $0, 1, 2 ..., E$}
    \FOR{$t$ = $0, 1, 2 ..., T$}
        \STATE With probability $\epsilon$ randomly choose an action $a_t$, \\               otherwise obtain $a_t=\mathrm{argmax}_{a}Q(s_t, a)$.
        \STATE Get context window sizes based on $a_t$ and sample the corresponding $x_t$ and $\hat{x}_{t}$ from both domains.
        \STATE Get occurrence of $a_t$ by performing $\mathcal{N}[a_t] += 1$.
        \STATE Generate the next state $s_{t+1}$ by feeding $x_t$ and $\hat{x}_t$ to the encoder $\mathcal{E}$.
        \STATE Compute $\mathcal{L}_{cls}$, $\mathcal{L}_{recon}$, $\mathcal{L}_{align}$ and $\mathcal{L}_{disc}$ with the generated feature vectors of two domains.
        \STATE Obtain $r_t$ via the reward function via Eq.~\ref{eq:reward}.
        \STATE Store the triplet $T_{t} = (s_{t}, a_{t}, s_{t+1}, r_{t})$ into $\mathcal{B}$
        \FOR{step = $1$, $2$, .., $S$}
            \STATE Optimize Q-function by sampling the data from $\mathcal{D}$ with Eq.~\ref{eq:qlearning}.
        \ENDFOR
    \ENDFOR
\ENDFOR
\end{algorithmic}
\end{algorithm}

\section{Related Works} 
\label{appendix:relate}
\subsection{Domain Adaptation}
Domain adaptation~\cite{wilson2020survey,tan2018survey} aims on transferring knowledge learned
from a source domain with label information to the target domain without sufficient labels. Maximum Mean Discrepancy (MMD) is a common method for reducing domain discrepancy within a metric space. 
Deep domain confusion~\cite{tzeng2014deep} minimizes the distance between the source and target distributions within a kernel-reproducing Hilbert space. Deep Adaptation Network~\cite{long2015learning} minimizes domain discrepancy in Hilbert space with an optimal multi-kernel selection method for matching domain embeddings. Another approach is domain discrimination, which is built upon adversarial learning framework~\cite{goodfellow2014generative} to generate domain-invariant features that cannot be discerned by a domain discriminator. Gradient reversal layer~\cite{ganin2015unsupervised} multiplies the gradient by a certain negative constant in the backpropagation stage to fool the domain discriminator and integrate model training with domain adaptation in a single process. Adversarial Discriminative Domain Adaptation~\cite{tzeng2017adversarial} proposes an unified framework for adversarial domain adaptation. We integrate MMD with domain discrimination and generate domain-invariant features to minimize domain discrepancies.
\vspace{-2pt}
\subsection{Time Series Domain Adaptation}
Although there are extensive studies on domain adaptation, most of them focus on applications of image data. Furthermore, due to the complex nature of time series data, performing domain adaptation on time series data not only requires considering instance-level correlation between two domains but also temporal dependencies across multiple data instances. To this end, Variational Recurrent Adversarial Deep Domain Adaptation~\cite{purushotham2016variational} is built upon Variational Recurrent Neural Network~\cite{chung2015recurrent} to capture temporal dependencies within multivariate time series data while generating domain-invariant features with adversarial objectives to perform domain adaptation. Domain Adversarial Neural Network~\cite{da2020remaining} intuitively adopts RNN based feature extractors to extract the representation of time series data and performs domain discrepancy minimization on the representations from different domains. However, the two methods align the two domains without considering the associative structure of time series variables. In other words, contextual differences between two domains are not taken into consideration during domain adaptation.

To alleviate the issue, Domain Adaptation Forecaster~\cite{jin2021attention} develops a dual encoder-decoder structure for two domains with self-attention layers and extracts domain-invariant features based on the attention weights from two domains. Sparse Associative Structure Alignment~\cite{cai2021time} further minimizes the discrepancies between the associative structure of time series variables from two domains and globally identifies a unified context window size in individual domains via a self-attention mechanism. However, all existing approaches are designed for time series regression and classification problems that focus on aligning the majority distribution of two domains and may potentially result in negative transfer on minority distribution. As transferring knowledge of anomalies between two domains requires aligning minority distribution of time series and the anomaly label spaces from two domains often share very limited similarity, all existing methods are unable to be adopted to solve the anomaly detection problem.
\vspace{-10pt}
\subsection{Partial Domain Adaptation}
One straightforward approach to transferring minority information is partial domain adaptation, which aims at developing instance-wise selective transfer strategies to alleviate the strong assumption of identical label space for different domains. To select the instance via class-level weighting, Selective Adversarial Network~\cite{cao2018partial} introduces multiple source discriminators to weigh the source samples according to the class probabilities predicted by the discriminators. To generate the weighting strategy by instance-level, Importance Weighted Adversarial Nets~\cite{zhang2018importance} derives the probability of a source example belonging to the target domain and weighs the source examples based on the probability for the domain-adversarial network. Example Transfer Network~\cite{cao2019learning} weighs the source samples according to the similarities measured by a domain discriminator and down-weights irrelevant source samples when updating the parameters of the source classifier.

As the key technique toward partial domain adaptation relies on iterative instance selection, recent advancements formulate the instance selection problem into a Markov decision process and adopt deep reinforcement learning algorithms to go through individual instances for partial domain adaptation. Domain Adversarial Reinforcement Learning~\cite{chen2020domain} iteratively selects instances from a candidate set to a finalized set for performing domain adaptation with a tailored reward function based on the domain adversarial learning framework. Reinforced Transfer network~\cite{chen2020selective} eliminates outlier samples in source classes through a reinforced data selector by considering both high-level and pixel-level information. However, instance-wise selective transfer methods are not applicable to time series data since there exist temporal dependencies between instances and arbitrary selection of partial instances may fail to consider the dependencies during domain adaptation. As a result, we propose a contextual domain adaptation framework that focuses on context sampling for individual instances in time series data.

\section{Dataset Description}
\label{appendix:data}
\begin{itemize}[leftmargin=15pt]
    \item \textbf{SMD} is a 5-week long dataset collected from a large Internet company. It records the server connection status. Individual servers are monitored by identical set of attributes (e.g., CPU mload, Disk write, TCP timeout), which made them applicable to the homogeneous setting. The task is to detect the unusual behaviors of the server connections.
    
    \item \textbf{Boiler} is collected from three sets of sensors that monitored three boilers with time span from 2014/3/24 to 2016/11/30. All boilers are the monitored by same type of sensors (e.g. outdoor temperature, gas volume, tube temperature), which lead to identical dimensions for all boilers and therefore applicable to the homogeneous setting. The task is to detect abnormal blow down of the boilers.
    
    \item \textbf{MSL and SMAP} are real spacecraft telemetry data collected from NASA. The two datasets are generated by two different telemetry tools (i.e., the Curiosity rover for MSL and observation satellite for SMAP) for two different planets (i.e., Mars for MSL and Earth for SMAP). The task for MSL dataset is to identify abnormal incidents on mars surface, where the task for SMAP dataset is to detect unusual surface activity on Earth. We adopt this dataset for heterogeneous transfer since the tasks of two datasets are similar, and the attributes are all about monitoring the surface of a planet.
\end{itemize}

We adopt the public available preprocess script provided by ~\cite{su2019robust}\footnote{\url{https://github.com/NetManAIOps/OmniAnomaly/blob/master/data_preprocess.py}} to preprocess SMAP, MSL and SMD datasets. For the Boiler dataset, we adopt the preprocessed dataset provided on the GitHub repository of~\cite{cai2021time}\footnote{https://github.com/DMIRLAB-Group/SASA/tree/main/datasets/Boiler}. The detail data statistics are tabulated in Table~\ref{tab:dataset}.
\begin{table}[!h]
    \scriptsize
    \centering
    \begin{tabular}{l|cccc}
    \toprule
    \textbf{Dataset} & \textbf{\# Timestamps} & \textbf{\# Entities} & \textbf{\# Dim.} & \textbf{\% Anomaly} \\ \midrule
    SMD & 708,405 & 8 & 25 & 13.1\% \\
    Boiler & 277,152 & 3 & 274 & 15.0\% \\
    SMAP & 429,735 & 1 & 55 & 4.1\% \\ 
    MSL & 66,709 & 1 & 38 & 10.7\% \\ 
    \bottomrule
    \end{tabular}
    \caption{Statistics of the datasets.}
    \label{tab:dataset}
\end{table}

\section{Evaluation Protocol}
\label{appendix:evaluate}
\begin{itemize}
    \item \textbf{Macro-averaged F1-score} calculates the F-measure separately for normal and anomaly classes, then two F-measure of two classes are averaged to be reported.
    \item  \textbf{AUC-ROC} leverages the produced anomalous score with continuous thresholding to create a series of points along ROC space and depicts the trade-of between correctly classified positive samples and incorrectly classified negative samples.
\end{itemize}
The intuition behind this is that the Macro-F1 ignores the imbalance between normal data points and outliers and therefore able to accurately reflect the prediction performance on both classes; while the AUC score reflects the quality of decision scores when predicting the labels with different thresholds. 

\section{Baselines} 
\label{appendix:baseline}
We compare the proposed ContexTDA with both single-domain anomaly detection models and dual-domain anomaly detector with different domain adaptation methods. All of the models are all trained and tested in unsupervised manner on the target domain. The single-domain baselines are trained without any information from the source domain, where the dual-domain models are trained with both label and data of the source domain.
\begin{itemize}
    \item \textbf{AE-MLP}~\cite{sakurada2014anomaly} is a a single-domain baseline, which is an 256-128-128-256 fully-connected autoencoder. AE-MLP is as a common baseline, which leverage reconstruction error on individual time points for anomaly detection.
    \item \textbf{AE-LSTM}~\cite{malhotra2016lstm} is a single-domain time series anomaly detection baseline, which built upon an autoencoder with 256-128-128-256 LSTM units. 
    \item \textbf{RDC} is built upon deep domain confusion~\cite{tzeng2014deep} and AE-LSTM. It aligns the two domains with MMD and leverages source domain labels to train the encoder of AE-LSTM for label prediction while minimizing the reconstruction error between the input data and the decoded output.
    \item \textbf{RDC-VRADA} combines deep domain confusion~\cite{tzeng2014deep} with variational recurrent adversarial deep domain adaptation~\cite{purushotham2016variational}, which simultaneously optimizes source domain label prediction, MMD and domain discrimination with the latent representations generated by the LSTM encoder. Meanwhile, the reconstruction objective of AE-LSTM is performed for detecting anomalies. It can be treated as the ContexTDA with a fixed context sampling strategy that constantly sample same context size.
    \item \textbf{SASA}~\cite{cai2021time} exploits self-attention layer with LSTM units to identify the optimal global context windows for source and target domain respectively and aligns the two domains with MMD. We directly use the class prediction probability to evaluate anomalous scores for each time point. 
    \item \textbf{RandContexTDA} is a simplified version of the proposed framework which randomly select context window sizes for each time point to examine the effectiveness of policy learning.
\end{itemize} 

\section{Implementation Details.} 
For \textbf{single-domain} baselines, we adopt the code of public available repository of AE-LSTM~\cite{malhotra2016lstm}\footnote{\url{https://github.com/PyLink88/Recurrent-Autoencoder}}; AE-MLP follows the implementation of PyOD~\footnote{https://github.com/yzhao062/pyod/}. The neural architectures for both baselines are $256-128-128-256$ with different types of neural units. For \textbf{dual-domain} baselines (RDC and RDC-VRADA), we follow the framework of public available GitHub repository~\footnote{https://github.com/syorami/DDC-transfer-learning} and modify the underlying neural architecture based on the aforementioned single-domain baselines for anomaly detection. As for SASA, we modify the implementation of public available GitHub repository~\footnote{https://github.com/DMIRLAB-Group/SASA} to unify the neural architecture of LSTM units and source domain classifier with our framework. We implement our anomaly detector based on the RDC-VRADA. Specifically, we implement two 128-128 MLP classifiers with dropout ratio $0.2$ and sigmoid activation function: one for performing source domain classification task with $\mathcal{L}_{cls}$, another one for generating domain invariant features with $\mathcal{L}_{disc}$. Then, we adopt the deep Q-learning implemented by the open-source package RLCard~\footnote{\url{https://bit.ly/3FVwO6S}} and use a 256-128-64 MLP for the Q-function, where the epsilon is set to $0.2$. The size of memory buffer is $10000$ and the sampling batch size for training DQN is set to $64$.
All algorithms are trained with epoch $10$, batch size $128$ where learning rates for each algorithm are chosen from \{0.05, 0.1, 0.15, 0.2, 0.25\} and the contamination ratios for all algorithms are chosen from \{0.01, 0.05, 0.1, 0.15, 0.2, 0.25, 0.3\}. For fixed window sized baseline (i.e., AE-MLP, AE-LSTM, RDC, RDC-VRADA), we empirically select context window size as $30$ for SMD dataset, $10$ for Boiler dataset and $40$ for MSL and SMAP datasets; and use the window size above as the maximum window size and create candidate window sizes from 1 to maximum window size for self-attention of SASA and action space of RandContexTDA and ContexTDA.

\end{document}